\documentclass[runningheads]{llncs}

\usepackage{accv}

\usepackage{accvabbrv}

\usepackage{graphicx}
\usepackage{booktabs}
\usepackage{multirow}
\usepackage{array}
\usepackage{algorithm}
\usepackage{algorithmic}
\usepackage{tikz}

\usepackage[accsupp]{axessibility}  
\usepackage{comment}

\usepackage{hyperref}

\usepackage{orcidlink}

\begin{document}

\title{WHERE to Generate Matters: Budget-Aware Synthetic Augmentation for Label Skewed Federated Learning}

\titlerunning{Budget-Aware Synthetic Augmentation for Label Skewed FL}

\author{Sangwoo Lee \and
Sunghwan Park\orcidlink{0000-0002-0253-110X} \and
Jaewoo Lee\orcidlink{0000-0001-5887-2184}}

\authorrunning{S.~Lee et al.}

\institute{Chung-Ang University, Seoul, South Korea
\email{\{dltkddn,tjdghks994,jaewoolee\}@cau.ac.kr}}

\maketitle

\begin{abstract}
Label skew in federated learning (FL) causes client drift and degrades global accuracy.
Synthetic data augmentation can reduce this imbalance; however, full class balancing requires substantial computation cost.
We propose FedEAS, a policy that assigns each client an entropy-adaptive per-class generation budget computed from its local label distribution.
The budget jointly decides \emph{how much} each client generates and \emph{WHERE} the samples go.
Accordingly, the total generation budget follows from the per-client budgets rather than being fixed in advance.
FedEAS recovers most of the accuracy gain of full class balancing while reducing the generation budget by 94.1\%.
At the same total generation budget, it outperforms Uniform allocation by up to 18.82\% across CIFAR-10 and CIFAR-100.

\end{abstract}

\section{Introduction}
\label{sec:intro}

Label skew is a fundamental challenge in federated learning (FL) \cite{kairouz,lifl}.
Clients hold locally imbalanced class distributions, so their local updates drift apart and degrade the global model \cite{zhao,liconv,hsu,noniidsilos}.
Optimization- and model-level methods mitigate this client drift with proximal regularization, control variates, contrastive objectives, and classifier calibration, leaving the local data distribution unchanged \cite{fedprox,scaffold,moon,classcal,nofear}.

Existing synthetic data augmentation methods correct the imbalance itself, generating the samples that each client's local distribution lacks \cite{genfedsd,feddda}.
The dominant policy is full class balancing, where every client generates until each of its classes reaches the size of its largest one \cite{genfedsd,feddda}.

\begin{figure}[!t]
\centering
\subfloat[Per-client generation demand]{%
  \includegraphics[width=0.5\textwidth]{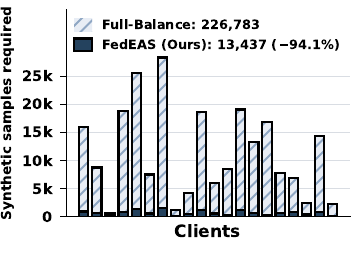}}\hfill
\subfloat[Allocation drives accuracy]{%
  \includegraphics[width=0.5\textwidth]{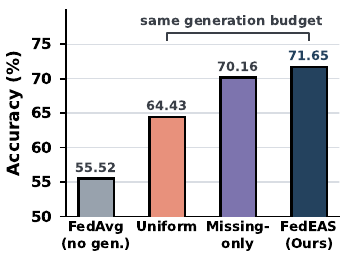}}
\caption{Budget-aware synthetic augmentation on a CIFAR-10 Dirichlet partition ($K=20$, $\alpha=0.1$). \textbf{(a)}~Reaching full class balancing requires 226{,}783 synthetic samples, whereas the FedEAS($\beta\!=\!12$) budgets require 13{,}437; hatched bars show per-client full-balancing demand, solid bars the FedEAS budgets. \textbf{(b)}~Allocation decides how much of the budget turns into accuracy. FedAvg trains without synthetic data, while Uniform, Missing-only, and FedEAS($\beta\!=\!12$) share the same 13{,}437-sample total and differ only in where the samples go (last-20-round mean).}
\label{fig:motivation}
\end{figure}

However, full class balancing is an expensive policy.
As shown in Fig.~\ref{fig:motivation}(a), reaching it on a 20-client CIFAR-10 partition under Dirichlet label skew with $\alpha=0.1$ requires 226{,}783 synthetic samples.
The cost stems from treating the amount to generate as a balancing requirement rather than a decision.
Recent studies relax this requirement by treating generation budget and client resources as explicit constraints \cite{fedgc,fimi}.
Their allocations are uniform or follow data size and device resources rather than label skew, and the total generation budget is fixed in advance.
As illustrated in Fig.~\ref{fig:motivation}(b), the same total generation budget yields very different accuracy depending on where the samples go.
Therefore, we ask the central question of this paper.
\emph{How much synthetic data does each client actually need, and WHERE should the samples go?}

To answer this question, we propose FedEAS (Federated Entropy-Adaptive Synthesis), which assigns each client a per-class generation budget computed from its local label distribution.
The budget jointly decides \emph{how much} each client generates and \emph{WHERE} the samples go.
A more skewed client receives a larger budget, a near-balanced client receives almost none, and in the IID (Independent and Identically Distributed) limit FedEAS reduces to standard FedAvg \cite{fedavg}.
Within each client, generation fills only the classes that fall below the budget.
These are exactly the client's absent and scarce classes.
Accordingly, the total generation budget follows from the per-client budgets rather than being fixed in advance.

On CIFAR-10 and CIFAR-100 under Dirichlet label skew, FedEAS recovers most of the accuracy gain of full class balancing while generating 94.1\% fewer synthetic samples (13{,}437 vs.\ 226{,}783 samples on CIFAR-10; Section~\ref{sec:main_acc}).
The saving translates into 3.7$\times$ lower end-to-end runtime on a single GPU (Section~\ref{sec:sensitivity}).
Accuracy also rises monotonically as the budget grows, so the operating point can be chosen by the available generation resources alone (Section~\ref{sec:sensitivity}).

This efficiency depends on \emph{WHERE} the samples go.
We therefore compare allocation policies under a matched-budget protocol, in which every policy shares the same generator and the same total generation budget, so that accuracy differences reflect allocation alone.
Under this protocol, FedEAS outperforms Uniform allocation by up to 18.82\% across CIFAR-10 and CIFAR-100, and reaches the 70\% accuracy threshold on CIFAR-10 up to $2\times$ faster than Missing-only generation (Sections~\ref{sec:matched} and~\ref{sec:sensitivity}).

In summary, this paper makes the following contributions.
\begin{itemize}
    \item \textbf{Entropy-adaptive budget allocation policy.} We propose FedEAS, which decides \emph{how much} and \emph{WHERE} to generate for reducing imbalance in label skewed FL. It gives skewed clients larger budgets and allocates samples only to each client's scarce classes. The total generation budget follows from the per-client budgets (Section~\ref{sec:method_alloc}).
    \item \textbf{Generation-efficient accuracy.} FedEAS recovers most of the accuracy gain of full class balancing with 94.1\% less generation budget and 3.7$\times$ lower end-to-end runtime (Sections~\ref{sec:main_acc} and~\ref{sec:sensitivity}).
    \item \textbf{Allocation dominance at matched budgets.} Under the same total generation budget, FedEAS outperforms Uniform allocation by up to 18.82\% and converges up to $2\times$ faster than Missing-only generation (Sections~\ref{sec:matched} and~\ref{sec:sensitivity}).
\end{itemize}

\section{Preliminaries and Motivation}

\subsection{Federated Setup and Client Drift}
\label{sec:prelim}

We consider a standard FL setup \cite{lifl} with $K$ clients, where client $k$ holds a local dataset $\mathcal{D}_k$ of size $N_k$. The global objective is
\begin{equation}
\min_{\theta \in \mathbb{R}^d} F(\theta) := \sum_{k=1}^{K} p_k F_k(\theta), \quad p_k = \frac{N_k}{\sum_j N_j},
\label{eq:objective}
\end{equation}
where $F_k(\theta) = \sum_{c=1}^{C} q_k^c \ell_k^c(\theta)$ denotes client $k$'s local objective over $C$ classes, $q_k^c = n_k^c / N_k$ denotes its class distribution with $n_k^c$ the number of class-$c$ samples, and $\ell_k^c$ denotes the class-$c$ loss.
Under non-IID data, FedAvg's convergence bound contains the cross-client heterogeneity term
\begin{equation}
\Gamma = \sum_k p_k \|\nabla F_k - \nabla F\|^2.
\label{eq:gamma}
\end{equation}
$\Gamma$ quantifies this heterogeneity in FedAvg convergence analysis \cite{scaffold,fedopt,wangfedavg} and is the starting point for our analysis (Section~\ref{sec:motivation}).

\textbf{Label skew and entropy.}
We induce label skew via Dirichlet partitioning \cite{hsu,noniidsilos} with concentration $\alpha$ (smaller $\alpha \Rightarrow$ more severe skew).
Client $k$'s skew is summarized by the normalized Shannon entropy
\begin{equation}
\tilde{H}_k = \frac{H_k}{\log C} \in [0, 1], \quad H_k = -\sum_{c=1}^{C} \frac{n_k^c}{N_k} \log \frac{n_k^c}{N_k},
\end{equation}
with $0 \log 0 := 0$, and the per-class mean count $\bar{n}_k = N_k / C$, used by the budget rule of Section~\ref{sec:method}.
Throughout, bars denote means ($\bar n_k$, $\bar q$) and tildes normalized or post-augmentation quantities ($\tilde H_k$, $\tilde q_k^c$).
The dominant policy is Full-Balance \cite{genfedsd}, the full class balancing of Section~\ref{sec:intro}, which augments until $\tilde q_k^c \approx 1/C$ for every $c$.
As shown in Fig.~\ref{fig:motivation}(a), it can require tens of thousands of samples for the most skewed clients, motivating a budget-aware reformulation.

\textbf{Research focus.}
We study how to allocate synthetic generation across clients and classes, not generator design.
All augmentation methods share one pre-trained class-conditional denoising diffusion probabilistic model (DDPM) \cite{ddpm,cfg,diffbeatgan} as a fixed black-box generator, so performance differences primarily reflect allocation.

\subsection{Motivation: How Much and WHERE to Generate}
\label{sec:motivation}

Section~\ref{sec:intro} asked \emph{how much} synthetic data each client needs and \emph{WHERE} the samples should go.
This subsection answers both parts before any training and arrives at three observations.
\textbf{Observation~1.} Imbalance is concentrated in a few clients.
\textbf{Observation~2.} Imbalance decreases only when samples go to scarce classes.
\textbf{Observation~3.} Imbalance becomes more expensive to remove as generation proceeds.
Observations~1 and~3 answer the \emph{how much} part, and Observation~2 answers the \emph{WHERE} part.
To verify them, we first build a score for allocations and show that it tracks the client drift of Section~\ref{sec:intro}.

\textbf{Scoring an allocation.}
We score an allocation by how strongly it pulls every client toward class balance.
For a total generation budget $B$, we define the imbalance score
\begin{equation}
\mathcal{I}(B) = \sum_{k=1}^{K} p_k \|q_k(B)-u\|_2^2,
\label{eq:surrogate}
\end{equation}
where $q_k(B)$ denotes client $k$'s class distribution after adding $B$ synthetic samples in total under a given allocation, $u=(1/C,\ldots,1/C)$ denotes the uniform distribution, and the weights $p_k$ keep their original values from Eq.~(\ref{eq:objective}).
$\mathcal{I}$ needs only label counts, so allocations can be compared before any generation.

\textbf{Why the score tracks client drift.}
$\mathcal{I}$ tracks the drift term $\Gamma$ of Eq.~(\ref{eq:gamma}).
To see the link, split $\mathcal{I}$ into cross-client heterogeneity and global imbalance:
\begin{equation}
\sum_k p_k\|q_k-u\|_2^2 = \sum_k p_k\|q_k-\bar{q}\|_2^2 + \|\bar{q}-u\|_2^2,
\label{eq:r-decomp}
\end{equation}
because $\sum_k p_k(q_k-\bar{q})=0$ with $\bar{q} := \sum_k p_k q_k$.
The first term is the distributional counterpart of $\Gamma$, and the second measures global deviation from balance, which matters for generalization on a balanced test set.
We reference $u$ rather than $\bar q$ so that the reference stays invariant under generation and $\mathcal{I}=0$ exactly when every client is balanced.
The link becomes explicit under three idealizing assumptions, namely (i) approximately orthogonal per-class loss gradients $\{\nabla\ell^c\}_c$, (ii) a shared gradient norm $\|g\|$, and (iii) per-class losses shared across clients ($\ell_k^c \equiv \ell^c$):
\begin{equation}
\Gamma \approx \|g\|^2 \sum_{k=1}^{K} p_k \|q_k-\bar{q}\|_2^2,
\label{eq:gamma-approx}
\end{equation}
in which the first term of Eq.~(\ref{eq:r-decomp}) appears exactly.
The derivation expands $\nabla F_k - \nabla F = \sum_c (q_k^c - \bar{q}^c)\nabla \ell^c$, where cross-terms vanish under (i) and diagonal terms share the scale $\|g\|^2$ under (ii). 
These assumptions do not hold tightly in deep networks, so we claim no formal bound, and the empirical counterpart $\hat\Gamma$ is measured directly in Section A of the supplementary material. 
Reducing class imbalance therefore reduces client drift at its distributional source, and $\mathcal{I}$ serves as its training-free surrogate.

\begin{figure}[!t]
\centering
\subfloat[Imbalance concentration]{%
  \includegraphics[width=0.5\textwidth]{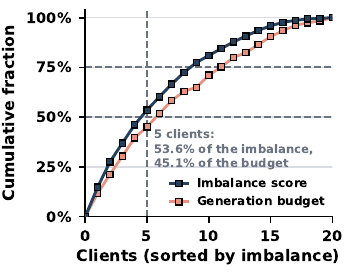}}\hfill
\subfloat[Imbalance reduction]{%
  \includegraphics[width=0.5\textwidth]{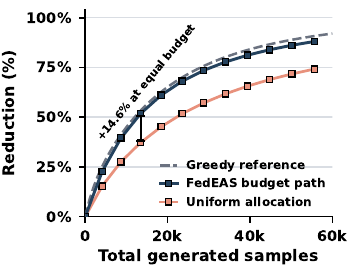}}
\caption{Imbalance concentration and reduction on the partition of Fig.~\ref{fig:motivation}. \textbf{(a)}~Cumulative fraction of the imbalance score and of the FedEAS generation budget ($\beta=12$), adding clients in decreasing order of imbalance; five clients hold 53.6\% of the imbalance score and receive 45.1\% of the budget. \textbf{(b)}~Normalized imbalance reduction $1-\mathcal{I}(B)/\mathcal{I}(0)$ as the total generation budget $B$ grows. Uniform allocation and the FedEAS budget path of Section~\ref{sec:method} are evaluated at the same matched budgets (markers), the dashed curve shows the greedy reference of Eq.~(\ref{eq:gain}), and the bracket marks the +14.6\% gap at the matched 13{,}437-sample budget.}
\label{fig:surrogate}
\end{figure}

\textbf{Observation 1: Imbalance is concentrated in a few clients.}
As shown in Fig.~\ref{fig:surrogate}(a), the top 25\% of clients hold 53.6\% of the total imbalance score on this CIFAR-10 partition ($K{=}20$, $\alpha{=}0.1$).
Accordingly, the generation required to fully balance a client varies by more than an order of magnitude, as illustrated in Fig.~\ref{fig:motivation}(a).
The normalized entropy $\tilde H_k$ captures this concentration from local counts alone.
A single balancing policy therefore conflates clients with very different needs.

\textbf{Observation 2: Imbalance decreases only when samples go to scarce classes.} Consider one synthetic sample of class $c$ added to client $k$. To first order, $\mathcal{I}$ changes by $-\Delta_k^c$ with
\begin{equation}
\Delta_{k}^{c} = \frac{2p_k}{N_k}\left(\|q_k\|_2^2 - q_k^c\right),
\label{eq:gain}
\end{equation}
which follows from the first-order form of the update $q_k \rightarrow (N_k q_k + e_c)/(N_k+1)$, where $e_c$ denotes the one-hot vector of class $c$.
The gain is positive only while $q_k^c$ lies below the client-specific level $\|q_k\|_2^2$, so a sample placed in an already-large class increases imbalance.
Uniform allocation pays for this directly.
On this partition it places 1{,}679 of its 13{,}437 samples, 12.5\%, at pairs where the gain is negative.
Since $p_k = N_k/\sum_j N_j$, the leading factor equals $2/\sum_j N_j$ for every client, and the gain grows with the client's skew, confirming Observation~1 at the sample level.
Efficient generation therefore fills scarce classes toward a per-client threshold instead of spreading over all classes.

\textbf{Observation 3: Imbalance becomes more expensive to remove as generation proceeds.}
Fig.~\ref{fig:surrogate}(b) traces the normalized imbalance reduction $1-\mathcal{I}(B)/\mathcal{I}(0)$ as the total generation budget grows, for uniform allocation and for a greedy reference that repeatedly places one sample at the pair maximizing Eq.~(\ref{eq:gain}).
Both curves are steepest at small budgets and flatten because the removable imbalance depletes.
A two-client case in Section B of the supplementary material shows the same diminishing-returns pattern in closed form.
The third curve in Fig.~\ref{fig:surrogate}(b) previews the policy of Section~\ref{sec:method}, which tracks the greedy reference using only local statistics and removes 51.8\% of the initial imbalance with 13{,}437 samples, 5.9\% of what Full-Balance requires.

Consequently, Observations~1--3 fix the requirements of an efficient policy.
It should give skewed clients larger budgets (Observation~1), allocate samples only to each client's scarce classes (Observation~2), and stop well before full class balancing (Observation~3).
Section~\ref{sec:method} meets all three requirements with a single rule.

\begin{figure}[!t]
\centering
\includegraphics[width=\textwidth]{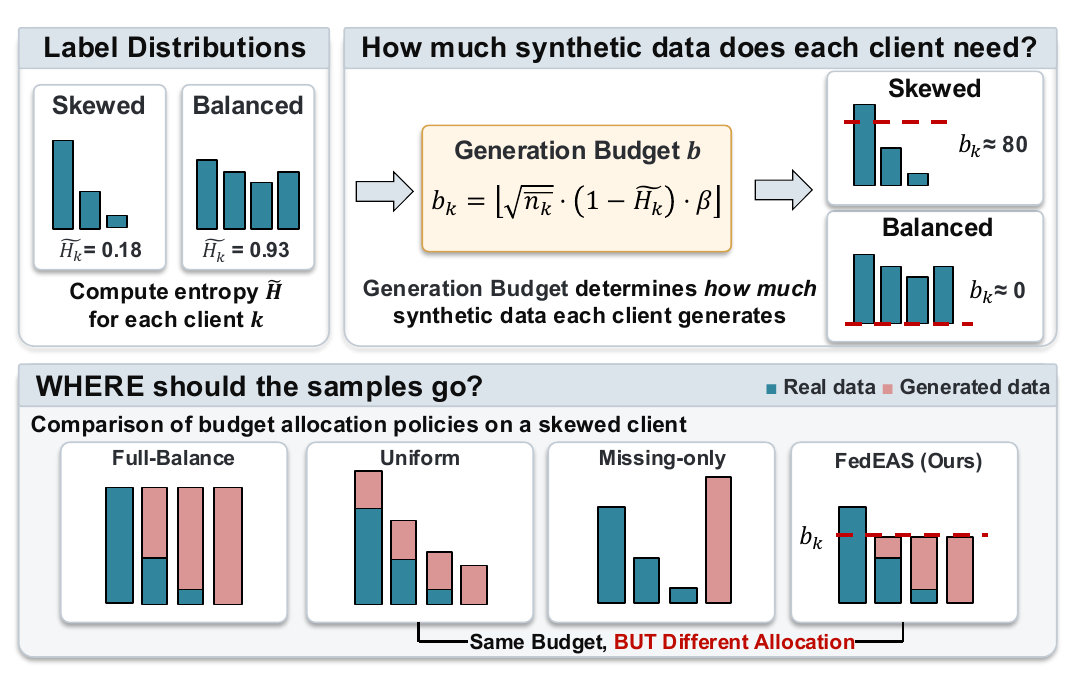}
\caption{Overview of FedEAS. Each client summarizes its skew by the normalized entropy $\tilde H_k$ (top left), computes the per-class generation budget $b_k$ of Eq.~(\ref{eq:cap}) (top right), and allocates samples only to the classes below $b_k$ (bottom). Full-Balance fills every class to its local maximum, while Uniform, Missing-only, and FedEAS place the same total generation budget differently, as compared in Section~\ref{sec:matched}.}
\label{fig:overview}
\end{figure}

\section{FedEAS: Federated Entropy-Adaptive Synthesis}
\label{sec:method}

FedEAS meets the three observations of Section~\ref{sec:motivation} with a single rule.
As illustrated in Fig.~\ref{fig:overview}, each client computes a per-class generation budget $b_k$ from its local label distribution, allocates samples only to the classes below $b_k$, and trains as in standard FedAvg.

\subsection{Entropy-Adaptive Budget Allocation Policy}
\label{sec:method_alloc}

FedEAS assigns client $k$ the per-class generation budget
\begin{equation}
b_k = \left\lfloor \sqrt{\bar{n}_k} \cdot (1 - \tilde{H}_k) \cdot \beta \right\rfloor,
\label{eq:cap}
\end{equation}
where $\bar{n}_k$ and $\tilde{H}_k$ are the per-class mean count and the normalized entropy of Section~\ref{sec:prelim}, and $\beta>0$ is the scalar budget parameter shared by all clients.
Here $b_k$ is a per-client, per-class count, and the floor makes it integer-valued.
Each factor of Eq.~(\ref{eq:cap}) answers one observation of Section~\ref{sec:motivation}.

\textbf{Skewed clients receive larger budgets (Observation 1).} 
The entropy factor $(1-\tilde{H}_k)$ grows with skew, so a more skewed client receives a larger budget and a near-balanced client receives almost none.
The $\sqrt{\bar{n}_k}$ factor scales the budget with client size sub-linearly.
A linear $\bar{n}_k$ rule would stretch the cross-client budget ratio to $38.5\times$, against $7.2\times$ under the sub-linear rule (CIFAR-10, $K{=}20$, $\alpha{=}0.1$), over-allocating to large clients while under-serving small ones.
In the IID limit, every $b_k$ reaches zero and FedEAS trains exactly as standard FedAvg.

\textbf{Samples go only to classes below the budget (Observation 2).}
Given $b_k$, class $c$ at client $k$ receives
\begin{equation}
n_{k,\mathrm{gen}}^c = \max\{0,\, b_k - n_k^c\},
\label{eq:gen}
\end{equation}
the fill-to-threshold form that Observation~2 derived.
Classes at or above $b_k$ receive nothing, so the samples concentrate on the client's absent and scarce classes.
Summing over clients and classes gives the realized total generation budget $B_{\mathrm{FE}}=\sum_{k,c} n_{k,\mathrm{gen}}^c$, the value of $B$ that Section~\ref{sec:motivation} scores and that every baseline shares under the matched-budget protocol of Section~\ref{sec:exp}.
If the pre-floor value of $b_k$ is below one, $b_k=0$ and client $k$ generates nothing.

\textbf{Early stopping of generation (Observation 3).}
The scalar $\beta$ sets the level of every budget, so generation ends in the steep region of Fig.~\ref{fig:surrogate}(b), while imbalance is still cheap to remove.
Raising $\beta$ scales every $b_k$ together, so a single parameter adapts the total generation budget to the available resources (Section~\ref{sec:sensitivity}).

\subsection{FedEAS in the Training Loop}

FedEAS changes only how each client builds its local dataset.
Algorithm~\ref{alg:fedeas} shows the full procedure.
The server broadcasts $\beta$ once at initialization.
On first participation, client $k$ computes Eqs.~(\ref{eq:cap})--(\ref{eq:gen}), generates the required images, and stores them in the cache $\mathcal{C}_k$.
The client then trains on $\mathcal{D}_k \cup \mathcal{C}_k$, and aggregation remains standard FedAvg weighted by the original dataset sizes $N_k$.

\textbf{One-time generation.}
Generation runs once per client and the cache is reused in all later rounds.
The saving over Full-Balance therefore comes from the smaller cache, and the communication protocol does not change.

\textbf{Generator-agnostic.}
Eqs.~(\ref{eq:cap})--(\ref{eq:gen}) read only local label counts.
Any class-conditional generator \cite{iddpm,scoresde,gan,stylegan2} can be substituted without changing the policy (Section~\ref{sec:gen_agnostic}).

\begin{algorithm}[!t]
\caption{FedEAS: entropy-adaptive budget allocation.}
\label{alg:fedeas}
\begin{algorithmic}[1]
\REQUIRE Clients $k=1,\ldots,K$; classes $c=1,\ldots,C$; local counts $n_k^c$ and $N_k=\sum_c n_k^c$; rounds $T$; local epochs $E$; budget parameter $\beta$; pre-trained class-conditional generator $G_\phi$.
\ENSURE Global model $\theta^T$.
\STATE Server initializes $\theta^0$ and broadcasts $\beta$; each client initializes synthetic cache $\mathcal{C}_k \leftarrow \emptyset$
\FOR{$t = 0, 1, \ldots, T-1$}
    \STATE Server broadcasts $\theta^t$ to selected clients $S_t$
    \FOR{each client $k \in S_t$ in parallel}
        \IF{$\mathcal{C}_k = \emptyset$ \textit{(one-time generation)}}
            \STATE $\bar{n}_k \leftarrow N_k/C$, $\tilde{H}_k \leftarrow H_k/\log C$
            \STATE $b_k \leftarrow \lfloor \sqrt{\bar{n}_k}\cdot(1-\tilde{H}_k)\cdot\beta \rfloor$
            \FOR{each class $c$ with $n_k^c < b_k$}
                \STATE $n_{k,\mathrm{gen}}^c \leftarrow b_k - n_k^c$
                \STATE Generate $n_{k,\mathrm{gen}}^c$ images of class $c$ via $G_\phi$
            \ENDFOR
            \STATE $\mathcal{C}_k \leftarrow$ all generated images
        \ENDIF
        \STATE Train on $\mathcal{D}_k \cup \mathcal{C}_k$ for $E$ local epochs; send $\theta^{t+1}_k$
    \ENDFOR
    \STATE $w_k \leftarrow N_k / \sum_{j \in S_t} N_j$ for each $k \in S_t$
    \STATE $\theta^{t+1} \leftarrow \sum_{k \in S_t} w_k \theta^{t+1}_k$
\ENDFOR
\end{algorithmic}
\end{algorithm}

\section{Experiments}
\label{sec:exp}

In this section, we validate the entropy-adaptive budget allocation policy on both parts of the central question of Section~\ref{sec:intro}.
Section~\ref{sec:main_acc} measures \emph{how much} generation budget FedEAS needs and isolates the effect of \emph{WHERE} the samples go under the matched-budget protocol.
Section~\ref{sec:sensitivity} sweeps the budget parameter, selects the default, and reports the end-to-end cost, and Section~\ref{sec:ablation} ablates the entropy and $\sqrt{\bar{n}_k}$ factors of the budget rule.

\subsection{Setup and Matched-Budget Protocol}

\textbf{Datasets and partitioning.}
We evaluate on CIFAR-10 ($C=10$) and CIFAR-100 ($C=100$) \cite{cifar}, inducing label skew via Dirichlet partitioning with concentration $\alpha$ (primary $\alpha=0.1$, with sweeps $\alpha \in \{0.05, 0.1, 0.3\}$ on CIFAR-10).
We report $K=20$ (50\% participation) and $K=100$ (10\% participation), each for $T=200$ rounds.
Accuracy, in this section, is the last-20-round mean, the mean global test accuracy over rounds 181--200.

\textbf{Model and generator.}
The local model is ResNet-18 \cite{resnet}, and each client trains $E=10$ local epochs per round with SGD (lr $1\!\times\!10^{-3}$) and batch size 64.
For the primary experiments we train a separate class-conditional DDPM \cite{ddpm,ddim} on each dataset's full training set, isolating allocation from generator design.
Uncurated samples from this DDPM appear in Section C of the supplementary material.
Section~\ref{sec:gen_agnostic} validates the allocation gap under a generator change with an off-the-shelf Stable Diffusion (SD-turbo) backbone \cite{sdturbo}, and deployment limitations are discussed in Section~\ref{sec:disc}.
Remaining hyperparameters appear in Section D of the supplementary material.

\textbf{Baselines and budget matching.}
We compare against six established FL baselines \cite{fedavg, fedprox, scaffold, moon, fedgen, fedmix} and three augmentation policies under the same DDPM backbone and schedule.
Uniform and Missing-only policies are compared under the matched-budget protocol.
Each shares the same generator and the same total generation budget $B_{\mathrm{FE}}$ realized by FedEAS (Section~\ref{sec:method_alloc}), so accuracy differences reflect allocation alone.

\emph{Uniform} assigns $\lfloor B_{\mathrm{FE}}/(KC) \rfloor$ samples to every client--class pair and distributes the remainder one at a time, spending exactly $B_{\mathrm{FE}}$ samples.
This is the equal-allocation rule that prior budget-constrained generation work recommends \cite{fedgc}.
\emph{Missing-only} applies the same rule over absent pairs ($n_k^c=0$) only.
\emph{Full-Balance} generates until each class reaches $\max_c n_k^c$, instantiating Gen-FedSD's policy \cite{genfedsd}.
It exceeds the shared budget and serves as an upper-bound reference rather than a matched competitor.
Unless stated otherwise, FedEAS uses $\beta=12$, selected in Section~\ref{sec:sensitivity}.
Samples are cached before a client's first update, and aggregation weights use the original local dataset sizes.

\subsection{Main Accuracy and Generation Budget}
\label{sec:main_acc}
\label{sec:matched}

\begin{table}[!t]
\centering
\caption{Accuracy (\%, last-20-round mean) across datasets, heterogeneity levels, and client scales. \textbf{Bold} and \underline{underline} indicate the best and second-best results, respectively, excluding Full-Balance. The last two rows report total generation budgets.}
\label{tab:main_acc}
\renewcommand{\arraystretch}{1.12}
\footnotesize
\setlength{\tabcolsep}{3pt}
\resizebox{\textwidth}{!}{%
\begin{tabular}{@{}l cccc cccc@{}}
\toprule
\multirow{3}{*}{\textbf{Method}} & \multicolumn{4}{c}{\textbf{$K=20$ (50\% participation)}} & \multicolumn{4}{c}{\textbf{$K=100$ (10\% participation)}} \\
\cmidrule(lr){2-5}\cmidrule(lr){6-9}
& \multicolumn{3}{c}{\textbf{CIFAR-10}} & \textbf{CIFAR-100} & \multicolumn{3}{c}{\textbf{CIFAR-10}} & \textbf{CIFAR-100} \\
\cmidrule(lr){2-4}\cmidrule(lr){5-5}\cmidrule(lr){6-8}\cmidrule(lr){9-9}
& $\alpha\!=\!0.05$ & $\alpha\!=\!0.1$ & $\alpha\!=\!0.3$ & $\alpha\!=\!0.1$ & $\alpha\!=\!0.05$ & $\alpha\!=\!0.1$ & $\alpha\!=\!0.3$ & $\alpha\!=\!0.1$ \\
\midrule
FedAvg \cite{fedavg}     & 39.16 & 55.52 & 67.47 & 42.83 & 43.41 & 50.18 & 65.11 & 41.11 \\
FedProx \cite{fedprox}   & 39.16 & 56.40 & 67.53 & 42.46 & 43.44 & 50.63 & 64.65 & 41.04 \\
SCAFFOLD \cite{scaffold} & 38.39 & 60.16 & \textbf{75.28} & 45.92 & 30.16 & 51.04 & 67.32 & 43.48 \\
MOON \cite{moon}         & 37.08 & 58.05 & 69.24 & 41.79 & 36.75 & 43.08 & 60.09 & 38.83 \\
FedGen \cite{fedgen}     & 32.53 & 52.07 & 67.84 & 37.87 & 39.06 & 47.49 & 65.00 & 38.38 \\
FedMix \cite{fedmix}     & 35.81 & 52.42 & 67.59 & 40.95 & 36.05 & 50.16 & 48.00 & 40.28 \\
Uniform (DDPM)           & 54.97 & 64.43 & 71.27 & 45.97 & 51.59 & 58.81 & 67.56 & 43.72 \\
Missing-only (DDPM)      & \underline{66.58} & \underline{70.16} & \underline{73.50} & \textbf{51.69} & \underline{65.37} & \underline{70.18} & \underline{71.87} & \underline{51.71} \\
Full-Balance (DDPM)      & ---   & 79.91 & ---   & 59.12 & ---   & 79.33 & ---   & 57.47 \\
\textbf{FedEAS} ($\beta\!=\!12$) & \textbf{68.25} & \textbf{71.65} & 73.08 & \underline{51.13} & \textbf{70.41} & \textbf{72.55} & \textbf{72.83} & \textbf{52.04} \\
\midrule
FedEAS total gen.        & 18{,}080 & 13{,}437 & 5{,}675 & 29{,}639 & 36{,}697 & 33{,}650 & 15{,}666 & 74{,}005 \\
Full-Balance total gen.  & ---      & 226{,}783 & ---   & 612{,}492      & ---      & 246{,}394 & ---      & 826{,}762 \\
\bottomrule
\end{tabular}}
\end{table}

FedEAS recovers most of the Full-Balance accuracy gain at a small fraction of its generation budget.
Table~\ref{tab:main_acc} reports the main comparison across datasets, heterogeneity levels, and client scales.
On CIFAR-10 ($\alpha=0.1$, $K=20$), FedEAS reaches 71.65\% against Full-Balance's 79.91\%, two thirds of the gain over FedAvg (55.52\%).
It generates 13{,}437 samples against 226{,}783, a 94.1\% reduction in generation budget.
The recovered portion of the gain grows with scale.
At $K=100$, FedEAS reaches 72.55\% against 79.33\%, three quarters of the gain over FedAvg (50.18\%).
On CIFAR-100 it recovers half of the gain at $K=20$ and two thirds at $K=100$.
Full-Balance remains the accuracy upper bound.

Table~\ref{tab:main_acc} also shows under which skew levels synthetic data augmentation is required.
Under severe CIFAR-10 skew ($\alpha\le0.1$), every reported augmentation policy, including Uniform, which ignores label skew, exceeds all six FL baselines.
At $\alpha=0.05$, $K=20$, Uniform reaches 54.97\% against 39.16\% for the strongest FL baseline.
On CIFAR-100, Missing-only and FedEAS exceed the strongest FL baseline (SCAFFOLD, 45.92\% at $K=20$) by more than 5\%, and Uniform matches it.
FedEAS itself improves over FedAvg by up to 29.09\% (CIFAR-10, $\alpha=0.05$, $K=20$) and stays competitive under mild skew ($\alpha=0.3$), where SCAFFOLD becomes strong.
Peak accuracies over all rounds appear in Section E of the supplementary material.

Under the matched-budget protocol, Uniform and FedEAS share the total generation budget, so their accuracy differences reflect allocation alone.
FedEAS outperforms Uniform at every $K=20$ matched budget, by +13.28\% at $\alpha=0.05$, +7.22\% at $\alpha=0.1$, and +1.81\% at $\alpha=0.3$ on CIFAR-10, and by +5.16\% on CIFAR-100.
At $K=100$ every gap widens, reaching +18.82\% at $\alpha=0.05$ on CIFAR-10 and +8.32\% on CIFAR-100.
The smaller gap at $\alpha=0.3$ is expected. Near-balanced clients leave less room for allocation, and restoring absent classes already captures much of the gain.

\begin{figure}[!t]
\centering
\includegraphics[width=\textwidth]{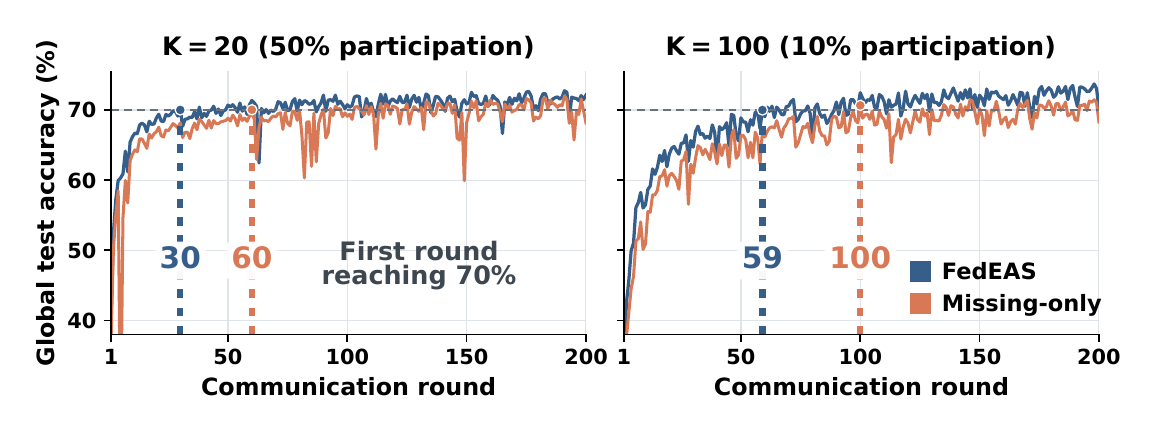}
\caption{FedEAS converges faster than Missing-only under matched budgets. The dashed line marks the 70\% global test accuracy threshold; dotted vertical lines mark the first round reaching it---30 vs.\ 60 rounds at $K=20$ and 59 vs.\ 100 at $K=100$.}
\label{fig:convergence_k20}
\label{fig:missingonly}
\end{figure}

\textbf{FedEAS converges faster than Missing-only at a matched budget.}
Missing-only restores absent classes but leaves scarce classes deficient, whereas FedEAS fills both toward the budget $b_k$.
The difference appears in convergence speed.
As shown in Fig.~\ref{fig:missingonly}, FedEAS reaches the 70\% global test accuracy threshold up to $2\times$ faster than Missing-only on CIFAR-10 ($\alpha=0.1$), with last-20-round means of 71.65 vs.\ 70.16 at $K=20$ and 72.55 vs.\ 70.18 at $K=100$.
The budget rule of Eq.~(\ref{eq:cap}) therefore accelerates convergence beyond Missing-only generation.

\textbf{The allocation gap holds under a generator change.}
\label{sec:gen_agnostic}
We repeat the matched-budget comparison with an off-the-shelf SD-turbo backbone \cite{sdturbo} at the matched 13{,}437-sample budget (CIFAR-10, $K=20$, $\alpha=0.1$).
The same pattern appears in peak accuracy.
FedEAS outperforms Uniform by 2.46\% under SD-turbo, against 6.34\% under the controlled DDPM, while absolute accuracy drops from generator and downsampling mismatch.

\begin{figure}[!t]
\centering
\subfloat[Accuracy at matched budgets]{%
  \includegraphics[width=0.5\textwidth]{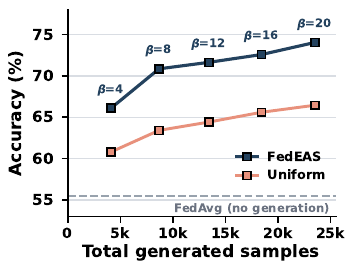}}\hfill
\subfloat[Budget parameter selection]{%
  \includegraphics[width=0.5\textwidth]{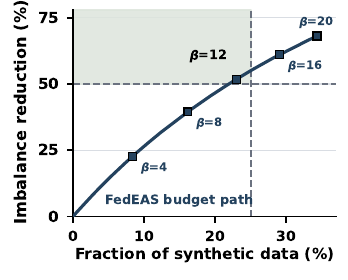}}
\caption{Budget sensitivity on CIFAR-10 ($\alpha=0.1$, $K=20$) under matched FedEAS--Uniform budgets. \textbf{(a)}~Last-20-round accuracy as the total generation budget grows. \textbf{(b)}~Selection of the default budget parameter. The budget path traces the imbalance reduction $1-\mathcal{I}(B)/\mathcal{I}(0)$ against the synthetic fraction of the training data as $\beta$ grows; markers are the sweep values. The dashed lines mark the two selection conditions. Only $\beta=12$ falls in the shaded region that removes at least 50\% of the initial imbalance while keeping synthetic samples below 25\% of the training data.}
\label{fig:results}
\end{figure}

\subsection{Budget Sensitivity and Selection}
\label{sec:sensitivity}

Accuracy rises monotonically as the budget grows.
We vary the budget parameter over $\beta\in\{4,8,12,16,20\}$ on CIFAR-10 ($K=20$, $\alpha=0.1$).
As shown in Fig.~\ref{fig:results}(a), the accuracy increases from 66.11\% to 74.05\%, exceeding Uniform by 5.30--7.59\% at every matched budget, so the operating point can be chosen by the available generation resources alone.

\textbf{The default budget parameter follows from two conditions.}
The realized budget should remove at least 50\% of the initial imbalance score $\mathcal{I}(0)$ of Section~\ref{sec:motivation}, and synthetic samples should stay below 25\% of the combined training data, so real samples outnumber synthetic ones at least three to one.
Both conditions read only label counts.
As shown in Fig.~\ref{fig:results}(b), the budget path satisfies both conditions only between $\beta\approx11.4$ and $\beta\approx13.4$ on this partition, and the sweep grid meets them only at $\beta=12$, which we adopt as the default.
At this operating point FedEAS removes 51.8\% of the initial imbalance while synthetic samples stay at 21.2\% of the combined training data.
A practitioner can re-run the same selection under different resource limits, and Fig.~\ref{fig:results}(a) shows the accuracy consequence of varying $\beta$.

\textbf{Efficiency falls as generation proceeds.}
We define the efficiency $\eta=\Delta\mathrm{Acc}\times10^3/\text{total generation budget}$,
where $\Delta\mathrm{Acc}$ is the accuracy gain over FedAvg (55.52\%).
Across this sweep, $\eta$ decreases monotonically from 2.575 to 0.788 while remaining
$1.7\times$--$2.0\times$ the Uniform efficiency at each matched budget, detailed in Section F of the supplementary material.
The imbalance that remains at larger budgets is more expensive to remove.
Full-Balance, the upper-bound reference, uses 226{,}783 samples and has $\eta=0.108$,
so full class balancing raises accuracy but lowers efficiency.

\textbf{Wall-clock cost.}
Under the same implementation on a single NVIDIA RTX 4090 GPU, FedEAS's runtime, including one-time generation and FL training, is 3h7m against 11h36m for Full-Balance.
The $3.7\times$ reduction shows the budget saving translates into measured end-to-end cost.

\subsection{Budget-Factor Ablation}
\label{sec:ablation}

The gains trace to allocation quality rather than to the budget level alone.
Table~\ref{tab:cap_ablation} compares FedEAS at $\beta=8$ with two variants on CIFAR-10 ($K=20$, $\alpha=0.1$), one without the entropy factor and one with the linear $\bar{n}_k$ rule.

FedEAS with both factors reaches 70.86\%, exceeding the variant without the entropy factor (70.34\%) and the linear-rule variant (70.33\%) despite generating fewer samples.
Removing the entropy factor lowers the last-20-round mean by 0.52\% while generating 1{,}664 additional samples.
The entropy factor therefore improves allocation rather than merely reducing the budget.

Replacing the $\sqrt{\bar n_k}$ factor with the linear $\bar{n}_k$ rule lowers the last-20-round mean by 0.53\% while generating 1{,}525 additional samples.
The linear rule also widens the cross-client budget ratio, as quantified in Section~\ref{sec:method_alloc}.
Together, these results support both factors of Eq.~(\ref{eq:cap}).

\begin{table}[!t]
\centering
\caption{Budget-factor ablation on CIFAR-10 ($K=20$, $\alpha=0.1$).}
\label{tab:cap_ablation}
\renewcommand{\arraystretch}{1.12}
\footnotesize
\setlength{\tabcolsep}{3.5pt}
\begin{tabular}{l c c c c}
\toprule
\textbf{Variant} & \textbf{Entropy} & \textbf{Size factor} & \textbf{Gen.} & \textbf{Accuracy} \\
\midrule
FedEAS ($\beta=8$) & yes & $\sqrt{\bar n_k}$ & 8,673 & \textbf{70.86} \\
w/o entropy & no & $\sqrt{\bar n_k}$ & 10,337 & 70.34 \\
Linear rule & yes & $\bar n_k$ & 10,198 & 70.33 \\
\bottomrule
\end{tabular}
\end{table}

\section{Related Work}

\textbf{Optimization- and model-level FL under label skew.}
Methods that do not alter local datasets intervene through the optimizer, representation, classifier, or loss \cite{feddf,fedrep,ditto,perfedavg}.
FedProx \cite{fedprox}, SCAFFOLD \cite{scaffold}, and MOON \cite{moon} use proximal regularization, control variates, and contrastive learning, respectively, while FedDyn \cite{feddyn} and FedNova \cite{fednova} correct client drift through dynamic regularization and normalized aggregation.
Label-skew-specific methods refine this via restricted softmax \cite{fedrs}, logits calibration \cite{fedlc}, classifier calibration \cite{classcal,ldam,decouple}, fixed classifiers \cite{nofear}, and label-distribution-aware concatenation \cite{diao}.
FedVLS \cite{guo} targets vacant classes via distillation and logit suppression at the model and loss level rather than by augmentation.
These are complementary to FedEAS, which intervenes at the data level before local training.

\textbf{Data-level augmentation in FL.}
Existing FL augmentation methods adapt classical augmentation \cite{augsurvey,cutmix,autoaugment,randaugment,cutout} but generally do not formulate augmentation as a client--class budget-allocation problem.
Gen-FedSD \cite{genfedsd} balances every client with text-to-image diffusion.
Our experiments evaluate this full class balancing as the upper-bound reference.
FedMix \cite{fedmix} shares averaged data to approximate cross-client Mixup \cite{mixup}, and FedGen \cite{fedgen} learns a data-free server-side generator over latent features.
FedFA \cite{fedfa} augments in feature space and FedRDN \cite{fedrdn} injects federation-wide statistics at the input level, both targeting feature rather than label skew, while Hong et al.\ \cite{hong} learn a shared augmentation policy.
Budget-constrained generation \cite{fedgc,fimi} fixes the total generation budget in advance and allocates it uniformly or by data size and device resources rather than by label skew.
FedEAS instead derives both the budget and its allocation from each client's label skew.
Generating only for absent pairs, our Missing-only baseline, is the closest policy but leaves scarce classes deficient (Section~\ref{sec:matched}).

\textbf{Centralized synthetic augmentation.}
Centralized studies show both the promise and the setup dependence of synthetic augmentation \cite{sariyildiz,stablerep}.
He et al.\ \cite{he} examine its benefits and shortcomings for recognition, Azizi et al.\ \cite{azizi} show ImageNet gains from diffusion samples, and Trabucco et al.\ \cite{trabucco} use diffusion edits for few-shot augmentation.
These studies motivate the budget-aware premise but not allocation across federated clients.

\section{Discussion}
\label{sec:disc}

\textbf{Limitations.}
Fixing a separate class-conditional DDPM \cite{ddpm} per dataset isolates allocation from generator design, but each DDPM is trained on its dataset's full training set, which limits deployment realism.
The SD-turbo comparison of Section~\ref{sec:gen_agnostic} partially decouples the allocation gap from the DDPM, yet privacy-preserving generator training remains open.
Related FL augmentation methods share this limitation \cite{genfedsd,fedgen,fedmix}.
The policy reads only local label counts, so a generator trained with privacy-preserving methods \cite{dpdiffusion,dpsgd,dpfl} can replace the DDPM.
DDPM sampling is also stochastic with no per-sample fidelity guarantee, which adds a generator-quality variance orthogonal to allocation.
Quality control such as FID filtering can be combined with the policy.

\textbf{Future work.}
Three directions extend this framework.
Federated-trained or stronger generators would address the deployment limitation above.
A tunable exponent $\bar{n}_k^\gamma$ would generalize the $\sqrt{\bar{n}_k}$ factor.
Budget allocation can also be paired with optimization-level corrections, and a preliminary FedEAS+SCAFFOLD run on CIFAR-10 ($\alpha=0.3$) reaches 78.18\%, exceeding either alone.

\section{Conclusion}

We recast synthetic augmentation for label skewed FL as a budget-allocation problem.
FedEAS addresses it with a single entropy-adaptive rule computed from local label counts.
Skewed clients receive larger budgets, samples go only to each client's scarce classes, and generation stops well before full class balancing.
The rule changes only the synthetic cache of each client, and the total generation budget follows from the per-client budgets.

FedEAS reduces the generation budget by 94.1\% and the end-to-end runtime by $3.7\times$ while recovering most of the accuracy gain of full class balancing.
With the total generation budget held equal, FedEAS outperforms Uniform allocation by up to 18.82\% on CIFAR-10 and CIFAR-100.
Allocation therefore decides how much of a fixed budget turns into accuracy, and \emph{WHERE} to generate is a primary design problem for synthetic augmentation in label skewed FL.

\newpage

%
%
\bibliographystyle{splncs04}
\bibliography{references}
\end{document}


\title{Supplementary Material\texorpdfstring{\\}{: }WHERE to Generate Matters: Budget-Aware Synthetic Augmentation for Label Skewed Federated Learning}
\titlerunning{Supplementary: Budget-Aware Synthetic Augmentation for Label Skewed FL}

\author{Sangwoo Lee \and
Sunghwan Park\orcidlink{0000-0002-0253-110X} \and
Jaewoo Lee\orcidlink{0000-0001-5887-2184}}

\authorrunning{S.~Lee et al.}

\institute{Chung-Ang University, Seoul, South Korea
\email{\{dltkddn,tjdghks994,jaewoolee\}@cau.ac.kr}}

\maketitle

\noindent This document provides supplementary material for the main paper. Equation, section, and table references of the form ``Eq.~(x)'', ``Section~x'' refer to the main paper.

\appendix

\section{Directional Gradient Alignment}
\label{app:gamma}

We measure a normalized empirical gradient-dissimilarity metric $\hat{\Gamma}$ under the main training schedule. In round $t$, each $k \in S_t$ completes $E=10$ epochs and produces a delta $\Delta_k^t = \theta^t - \theta_k^{t+1}$; normalizing as $\hat{\Delta}_k^t = \Delta_k^t/\|\Delta_k^t\|_2$ with weights $w_k^t = N_k/\sum_{j\in S_t}N_j$, we compute
\begin{equation}
\hat{\Gamma}_t = \sum_{k \in S_t} w_k^t \|\hat{\Delta}_k^t - \bar{\hat{\Delta}}^t\|^2,
\quad
\bar{\hat{\Delta}}^t = \sum_{k \in S_t} w_k^t \hat{\Delta}_k^t .
\end{equation}
We report the median of $\hat{\Gamma}_t$ over all $T=200$ rounds at 50\% participation; since its scale depends on sampling ratio and local epochs, we use it only as a controlled measurement under fixed participation and $E$.

\begin{table}[!ht]
\centering
\caption{Empirical gradient dissimilarity $\hat{\Gamma}$ on CIFAR-10 ($\alpha=0.1$, $\beta=12$), median over 200 rounds with 50\% client participation and $E=10$ local epochs. Uniform and FedEAS share the same total generation budget; Full-Balance is the upper-bound reference.}
\renewcommand{\arraystretch}{1.15}
\footnotesize
\setlength{\tabcolsep}{3pt}
\begin{tabular}{l c c c c}
\toprule
 & \textbf{FedAvg} & \textbf{Uniform} & \textbf{FedEAS} & \textbf{Full-Bal.} \\
\midrule
Last-20 avg.\ (\%) & 55.52 & 64.43 & 71.65 & 79.91 \\
Total gen.        & 0     & 13,437 & 13,437 & 226,783 \\
$\hat{\Gamma}$ (med.) & $1.42\!\times\!10^{-4}$ & $2.15\!\times\!10^{-5}$ & $1.95\!\times\!10^{-5}$ & $1.14\!\times\!10^{-6}$ \\
$\hat{\Gamma}$ reduction & --- & $84.9\%$ & $86.3\%$ & $99.2\%$ \\
\bottomrule
\end{tabular}
\end{table}

This analysis clarifies the mechanism. Because the $\hat{\Delta}_k^t$ are unit vectors, $\hat{\Gamma}_t = 1-\|\bar{\hat{\Delta}}^t\|^2$ measures only the dispersion of update directions, not which classes the consensus direction serves. Both matched-budget policies nearly eliminate directional disagreement (84.9\% vs.\ 86.3\% reduction), so alignment alone cannot account for their 7.22\% accuracy gap. This is consistent with prior reports that gradient dissimilarity is a loose predictor of practical FedAvg performance \cite{wangfedavg}. The gap instead traces to coverage. Uniform grants every client--class pair only $\lfloor B_{\mathrm{FE}}/(KC)\rfloor{=}67$ samples, too few for the scarcest pairs, whereas FedEAS fills each scarce class to the per-class generation budget $b_k$ of Eq.~(\ref{eq:cap}). Accordingly, the advantage of FedEAS widens in the most skewed configurations (Section~\ref{app:peak}).

\section{Supplementary Closed-Form Perspective}
\label{app:proof}

The main text uses the imbalance score $\mathcal{I}$ of Eq.~(\ref{eq:surrogate}) as the primary motivation because it applies directly to the multi-client, multi-class budget-allocation setting. For completeness, we also include a minimal two-client/two-class case in which the diminishing-returns reduction pattern can be written in closed form.

\begin{theorem}[Diminishing Returns of Generation]
\label{thm:dr}
Consider two clients with $C=2$ classes. Client~1 holds $n$ samples of class~A and 0 of class~B; client~2 holds the reverse. Adding $m$ synthetic samples of the deficient class to each client yields generation ratio $r=m/n$. Then the post-augmentation heterogeneity satisfies
\begin{equation}
\tilde{\Gamma}(r) = \Gamma_0 \cdot h(r), \quad h(r)=\left(\frac{1-r}{1+r}\right)^2,
\end{equation}
where $h(r)$ is strictly decreasing and strictly convex on $[0,1]$, with marginal reduction $|h'(r)|$ diminishing monotonically from 4 at $r=0$ to 0 at $r=1$.
\end{theorem}

\textbf{Proof.}
\textbf{Setup.} Client~1 holds $n$ samples of class~A and 0 of class~B; client~2 holds the reverse. Thus $q_1=(1,0)$, $q_2=(0,1)$, and with equal client weights $p_1=p_2=1/2$, the global class distribution is $\bar{q}=(1/2,1/2)$.

\textbf{Gradient deviation before augmentation.} Under cross-entropy loss with shared $\theta$, the expected client gradient is $\nabla F_k(\theta) = \sum_c q_k^c \nabla \ell^c(\theta)$. Defining $\delta^c(\theta) := \nabla \ell^c(\theta) - \nabla F(\theta)$,
\begin{equation}
\nabla F_k - \nabla F = \sum_c (q_k^c - \bar{q}^c) \delta^c.
\end{equation}
With $\delta^A = -\delta^B =: \delta$, client~1's deviation is $(1/2, -1/2)$, giving $\nabla F_1 - \nabla F = \delta$. By symmetry, $\nabla F_2 - \nabla F = -\delta$, hence $\Gamma_0 = \|\delta\|^2$.

\textbf{After augmentation.} With $r = m/n$, the augmented distribution for client~1 is $\tilde{q}_1 = (n/(n+m), m/(n+m))$, so $\tilde{q}_1^A - 1/2 = (1-r)/(2(1+r))$. Substituting,
\begin{equation}
\tilde{\Gamma}(r) = \left(\frac{1-r}{1+r}\right)^2 \|\delta\|^2 = \Gamma_0 \cdot h(r).
\end{equation}

\textbf{Monotonicity and convexity.} $h'(r) = -4(1-r)/(1+r)^3 < 0$ and $h''(r) = 8(2-r)/(1+r)^4 > 0$ for $r \in [0,1)$, so $h$ is strictly decreasing and strictly convex; $|h'(r)|$ decreases monotonically from 4 at $r=0$ to 0 at $r=1$. \hfill$\blacksquare$

\begin{figure}[!t]
\centering
\includegraphics[width=0.86\linewidth]
{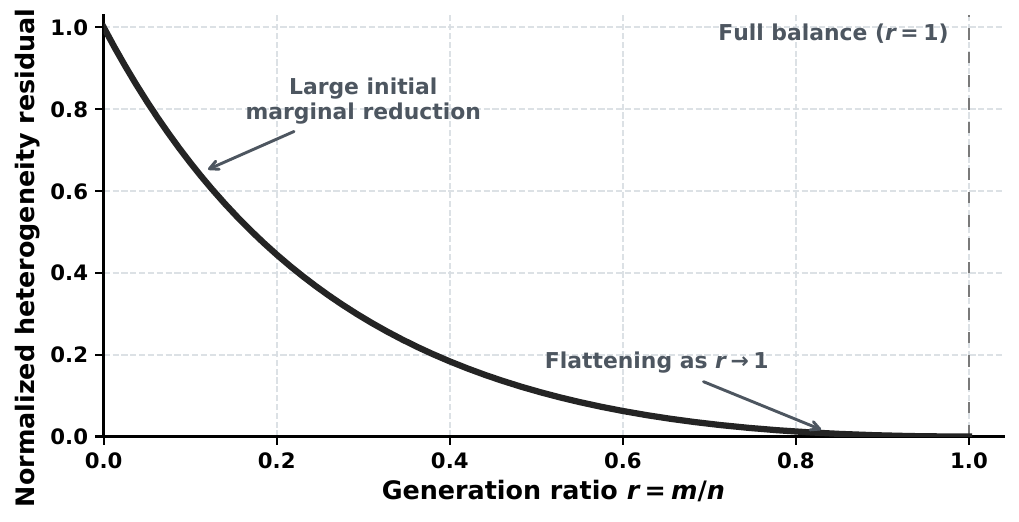}
\caption{
Closed-form normalized heterogeneity residual
$h(r)=((1-r)/(1+r))^2$ from Theorem~\ref{thm:dr}.
The residual decreases rapidly at small generation ratios and
flattens toward full balance ($r=1$), illustrating the diminishing
marginal reduction from additional generation.
}
\label{fig:theorem_diminishing_returns}
\end{figure}

\subsection{Extension to $K$ Clients and $C$ Classes (Sketch)}
\label{app:kclient}

Continuing the closed-form perspective above, we sketch how the qualitative principle can extend to the general $K$-client/$C$-class setting. Diminishing returns under a per-sample generation cost motivate an operating point below full balance. The goal is to support the plausibility of such an interior cost-aware solution under a surrogate objective, not to derive the FedEAS rule itself.

Define the per-pair benefit $g(r) := 1 - h(r) = 4r/(1+r)^2$. Since $g'(r) > 0$ and $g''(r) < 0$ on $[0,1)$, $g$ is strictly increasing and strictly concave. This concavity is the structural building block of the argument.

Consider the separable surrogate
\begin{equation}
J(\mathbf{m}) = \sum_{k} \sum_{c \in \mathcal{C}_{k,\mathrm{mean}}^-} a_k^c g_k^c(m_k^c) - \lambda \sum_{k} \sum_{c \in \mathcal{C}_{k,\mathrm{mean}}^-} m_k^c,
\end{equation}
where $m_k^c \geq 0$ is the synthetic count for $(k,c)$, $\mathcal{C}_{k,\mathrm{mean}}^-=\{c: n_k^c < \bar{n}_k\}$ is the set of classes deficient relative to the per-client mean, $a_k^c > 0$ is a pair weight, and $\lambda>0$ is a per-sample cost. Each $g_k^c$ is strictly concave (mirroring the 2-client/2-class case).

\textbf{Proposition.}
Let $d_k^c := \bar{n}_k - n_k^c > 0$, and suppose that
$g_k^c$ is differentiable and strictly concave on $[0,d_k^c]$.
If
\begin{equation}
a_k^c (g_k^c)'(0)
>
\lambda
>
a_k^c (g_k^c)'(d_k^c),
\end{equation}
then the unique maximizer of
\[
J_k^c(m)
=
a_k^c g_k^c(m)-\lambda m,
\qquad 0\leq m\leq d_k^c,
\]
satisfies
\[
0 < (m_k^c)^\star < d_k^c.
\]

\textit{Sketch.}
Strict concavity of $g_k^c$ implies that $J_k^c$ is strictly
concave and that $(J_k^c)'(m)$ is strictly decreasing.
The assumed inequalities give
\[
(J_k^c)'(0)>0,
\qquad
(J_k^c)'(d_k^c)<0.
\]
Therefore, $(J_k^c)'$ has a unique zero in $(0,d_k^c)$, which is
the unique maximizer. Thus the cost-aware solution generates a
positive amount while remaining strictly below full balance.
\hfill$\blacksquare$

Under this surrogate, the argument supports the existence of a cost-aware allocation below full balance but does not identify its location. FedEAS instantiates this principle through the entropy-adaptive, sub-linear rule of Eq.~(\ref{eq:cap}), validated empirically in Section~\ref{sec:exp}, where FedEAS uses 5.9\% of the Full-Balance generation budget on CIFAR-10 at $\alpha=0.1$.

\section{Uncurated DDPM Samples}
\label{app:samples}

\begin{figure}[H]
  \centering
  \begin{tikzpicture}
    \node[anchor=south west,inner sep=0] (img) at (0,0)
      {\includegraphics[width=0.74\linewidth]{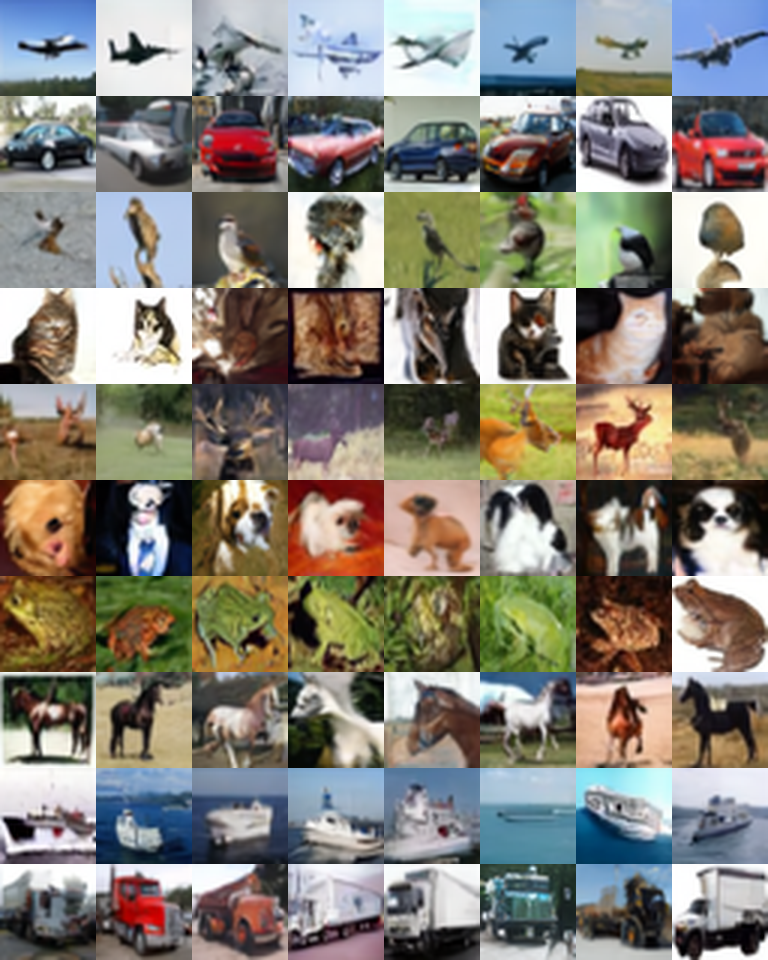}};
    \begin{scope}[x={(img.south east)},y={(img.north west)}]
      \foreach [count=\i] \name in {airplane,automobile,bird,cat,deer,dog,frog,horse,ship,truck}{
        \node[anchor=east,font=\small] at (-0.01,{1-(\i-0.5)/10}) {\name};
      }
    \end{scope}
  \end{tikzpicture}
  \caption{Uncurated random samples from the class-conditional DDPM (native $32\!\times\!32$, shown enlarged with nearest-neighbor interpolation for visibility) used by all DDPM-based augmentation policies. Rows correspond to the ten CIFAR-10 classes. Samples are provided for transparency; this work focuses on allocation rather than per-sample fidelity, and generator-quality variance is orthogonal to the allocation policy (Section~\ref{sec:disc}).}
  \label{fig:ddpm_samples}
\end{figure}

Fig.~\ref{fig:ddpm_samples} shows uncurated random samples from the class-conditional DDPM at CIFAR's native $32\!\times\!32$ resolution. The samples are class-recognizable but exhibit the limited per-sample fidelity expected at this resolution. As discussed in Section~\ref{sec:disc}, this is a generator-quality factor orthogonal to the allocation policy, since the policy reads only local label counts. Crucially, all augmentation policies (Uniform, Missing-only, Full-Balance, and FedEAS) draw from this same generator, so the visualized fidelity is held constant across policies and cannot account for their relative differences.

\section{Implementation Details}
\label{app:impl}

\textbf{Baseline hyperparameters.} FedProx: $\mu = 0.01$. SCAFFOLD: global lr $= 1.0$. MOON: $\tau = 0.5$, $\mu = 5$. FedGen: ensemble lr $= 3\!\times\!10^{-4}$, gen batch size $= 32$, noise/hidden dim $= 128/256$, ensemble epoch $= 100$, train generator epoch $= 10$. FedMix: $M_b$ and $\lambda$ following the original CIFAR-10 setup of \cite{fedmix}.

\textbf{DDPM training.} Architecture with \texttt{dim\_mults} $= (1,2,4,8)$, diffusion timesteps $= 1000$, batch size 128, 300{,}000 training steps, separate model per dataset on the full training set ($32\!\times\!32$ resolution). Sampling uses 250 timesteps with class conditioning.

\textbf{Hardware and runtime.} Experiments and wall-clock comparisons were run on a single NVIDIA RTX 4090 GPU. Reported runtimes include the one-time synthetic generation step and the subsequent FL training loop under the same implementation.

\textbf{Cache training protocol.} During local training, each client constructs $\mathcal{D}_k \cup \mathcal{C}_k$ via simple concatenation and trains on the full set with uniform random shuffling each epoch. No explicit ratio control is applied; the per-batch composition is determined by the natural ratio $|\mathcal{D}_k| : |\mathcal{C}_k|$. On CIFAR-10 at $\alpha=0.1$ with $\beta=12$, the median client holds $|\mathcal{D}_k| \approx 1{,}624$ real samples and $|\mathcal{C}_k| \approx 595$ synthetic samples (real fraction $\approx 73\%$).

\textbf{Combining FedEAS with optimization-based methods.} A preliminary combination of FedEAS ($\beta=12$) with SCAFFOLD on CIFAR-10 at $\alpha=0.3$ yields 78.18\%, exceeding both individual methods (FedEAS: 73.66\%, SCAFFOLD: 75.76\% under matched full-epoch training), suggesting that data-level and optimization-level corrections are complementary.

\textbf{Stable Diffusion validation details.} For the SD-based validation in Section~\ref{sec:gen_agnostic}, $512\!\times\!512$ SD-turbo generations are downsampled to $32\!\times\!32$ with anti-aliased resampling before being cached and reused under the same local training protocol as the DDPM-generated samples.

\section{Peak Accuracy Across Datasets, Heterogeneity, and Client Scale}
\label{app:peak}

The main paper reports the last-20-round mean to reflect converged performance. For completeness, Supp. Table~\ref{app:peak_acc} reports the peak accuracy, the best global-model accuracy attained at any of the 200 communication rounds, across both client scales ($K\!=\!20$ at 50\% participation and $K\!=\!100$ at 10\% participation), both datasets (CIFAR-10 and CIFAR-100), and three Dirichlet heterogeneity levels ($\alpha\in\{0.05,0.1,0.3\}$). The bottom rows list the total number of generated samples for FedEAS ($\beta\!=\!12$) and for the Full-Balance upper-bound reference, so that accuracy can be read against generation cost.

The peak-accuracy view corroborates the main results. At matched budgets, FedEAS attains the highest accuracy among the budgeted policies in most configurations, and its advantage over Uniform and Missing-only widens as heterogeneity increases (smaller $\alpha$) and as the client scale grows to $K\!=\!100$, where per-client label skew is most severe. Full-Balance remains the upper-bound reference but requires $226{,}783$ samples against $13{,}437$ on CIFAR-10 at $\alpha\!=\!0.1$, $K\!=\!20$, confirming that FedEAS recovers most of the attainable gain with a small fraction of the generation budget.

\begin{table}[H]
\centering
\caption{Peak accuracy (\%) over all 200 rounds across datasets, heterogeneity levels, and client scales. \textbf{Bold} indicates FedEAS results. Uniform and Missing-only share the FedEAS total generation budget; Full-Balance is the upper-bound reference. The bottom rows report total generated samples for FedEAS ($\beta\!=\!12$) and Full-Balance.}
\label{app:peak_acc}
\renewcommand{\arraystretch}{1.12}
\footnotesize
\setlength{\tabcolsep}{3pt}
\resizebox{\textwidth}{!}{%
\begin{tabular}{@{}l c c c c c c c c@{}}
\toprule
\multirow{3}{*}{\textbf{Method}} & \multicolumn{4}{c}{\textbf{$K=20$ (50\% participation)}} & \multicolumn{4}{c}{\textbf{$K=100$ (10\% participation)}} \\
\cmidrule(lr){2-5}\cmidrule(lr){6-9}
& \multicolumn{3}{c}{\textbf{CIFAR-10}} & \textbf{CIFAR-100} & \multicolumn{3}{c}{\textbf{CIFAR-10}} & \textbf{CIFAR-100} \\
\cmidrule(lr){2-4}\cmidrule(lr){5-5}\cmidrule(lr){6-8}\cmidrule(lr){9-9}
& $\alpha\!=\!0.05$ & $\alpha\!=\!0.1$ & $\alpha\!=\!0.3$ & $\alpha\!=\!0.1$ & $\alpha\!=\!0.05$ & $\alpha\!=\!0.1$ & $\alpha\!=\!0.3$ & $\alpha\!=\!0.1$ \\
\midrule
FedAvg \cite{fedavg}     & 44.92 & 61.11 & 68.93 & 43.44 & 50.81 & 56.08 & 67.51 & 41.76 \\
FedProx \cite{fedprox}   & 44.99 & 60.67 & 68.80 & 42.91 & 50.73 & 56.35 & 66.82 & 41.58 \\
SCAFFOLD \cite{scaffold} & 41.25 & 64.04 & 75.76 & 46.32 & 36.13 & 54.19 & 68.02 & 44.02 \\
MOON \cite{moon}         & 45.51 & 61.84 & 71.09 & 42.31 & 44.35 & 50.36 & 63.10 & 39.44 \\
FedGen \cite{fedgen}     & 39.54 & 60.24 & 70.55 & 38.69 & 47.69 & 53.99 & 68.47 & 38.94 \\
FedMix \cite{fedmix}     & 40.71 & 59.38 & 68.97 & 41.55 & 41.73 & 55.67 & 61.21 & 40.88 \\
Uniform (DDPM)           & 59.67 & 66.40 & 73.06 & 46.47 & 62.21 & 62.91 & 68.86 & 44.21 \\
Missing-only (DDPM)      & 69.00 & 72.00 & 74.43 & 52.61 & 70.11 & 71.46 & 72.70 & 52.14 \\
Full-Balance (DDPM)      & ---   & 80.28 & ---   & 59.39 & ---   & 79.79 & ---   & 57.75 \\
\textbf{FedEAS} ($\beta\!=\!12$) & \textbf{70.25} & \textbf{72.74} & \textbf{73.66} & \textbf{51.44} & \textbf{73.25} & \textbf{73.71} & \textbf{73.33} & \textbf{52.56} \\
\midrule
FedEAS total gen.        & 18{,}080 & 13{,}437 & 5{,}675 & 29{,}639 & 36{,}697 & 33{,}650 & 15{,}666 & 74{,}005 \\
Full-Balance total gen.  & ---      & 226{,}783 & ---   & 612{,}492      & ---      & 246{,}394 & ---      & 826{,}762 \\
\bottomrule
\end{tabular}}
\end{table}

\section{Full $\beta$-Sweep with Efficiency}
\label{app:beta}

Supp. Table~\ref{tab:beta} reports the complete budget sweep using the same last-20-round mean as the main comparison, adding the per-configuration efficiency metric $\eta = \Delta\text{FedAvg (\%)} \times 10^{3}/\text{Total gen.}$ (higher is better). FedEAS and Uniform are matched in total generation at each $\beta$; Full-Balance is the upper-bound reference. FedEAS is $1.7\times$--$2.0\times$ more efficient than Uniform at every matched budget, and efficiency decreases monotonically as $\beta$ grows, consistent with the diminishing-returns trade-off discussed in Section~\ref{sec:sensitivity}.

\begin{table}[H]
\centering
\caption{$\beta$ sweep on CIFAR-10 ($\alpha=0.1$), reported as the last-20-round mean (rounds 181--200). FedEAS and Uniform are matched at each $\beta$; Full-Balance is the upper-bound reference. Efficiency $\eta = \Delta\text{FedAvg (\%)} \times 10^{3}/\text{Total gen.}$; higher is better.}
\label{tab:beta}
\renewcommand{\arraystretch}{1.1}
\footnotesize
\begin{tabular}{l c c c c}
\toprule
\textbf{Method} & \textbf{Last-20 avg.} & \textbf{Total gen.} & \boldmath$\Delta$\textbf{FedAvg} & \boldmath$\eta$ \\
\midrule
FedAvg & 55.52 & 0 & --- & --- \\
FedEAS $\beta\!=\!4$  & 66.11 & 4{,}113  & +10.59 & 2.575 \\
FedEAS $\beta\!=\!8$  & 70.86 & 8{,}673  & +15.34 & 1.769 \\
FedEAS $\beta\!=\!12$ & 71.65 & 13{,}437 & +16.13 & 1.200 \\
FedEAS $\beta\!=\!16$ & 72.59 & 18{,}435 & +17.07 & 0.926 \\
FedEAS $\beta\!=\!20$ & 74.05 & 23{,}514 & +18.53 & 0.788 \\
\midrule
Uniform $\beta\!=\!4$  & 60.81 & 4{,}113  & +5.29  & 1.286 \\
Uniform $\beta\!=\!8$  & 63.42 & 8{,}673  & +7.90  & 0.911 \\
Uniform $\beta\!=\!12$ & 64.43 & 13{,}437 & +8.91  & 0.663 \\
Uniform $\beta\!=\!16$ & 65.61 & 18{,}435 & +10.09 & 0.547 \\
Uniform $\beta\!=\!20$ & 66.46 & 23{,}514 & +10.94 & 0.465 \\
\midrule
Full-Balance & 79.91 &  226{,}783 & +24.39 & 0.108 \\
\bottomrule
\end{tabular}
\end{table}

\bibliographystyle{splncs04}
\bibliography{references}